# AN ONTOLOGICAL ARCHITECTURE FOR ORBITAL DEBRIS DATA

Robert J. Rovetto
New York, USA
rrovetto@terpalum.umd.edu
ontologos@yahoo.com

**Abstract**. The orbital debris problem presents an opportunity for inter-agency and international cooperation toward the mutually beneficial goals of debris prevention, mitigation, remediation, and improved space situational awareness (SSA). Achieving these goals requires sharing orbital debris and other SSA data. Toward this, I present an ontological architecture for the orbital debris domain, taking steps in the creation of an orbital debris ontology (ODO). The purpose of this ontological system is to (I) represent general orbital debris and SSA domain knowledge, (II) structure, and standardize where needed, orbital data and terminology, and (III) foster semantic interoperability and data-sharing. In doing so I hope to (IV) contribute to solving the orbital debris problem, improving peaceful global SSA, and ensuring safe space travel for future generations.

## SYNOPSIS / EXTENDED ABSTRACT

The orbital debris problem presents an opportunity for inter-agency and international cooperation toward the mutually beneficial goals of:

- Debris detection, identification, tracking and orbit propagation
- Orbital debris prevention, mitigation and remediation
- Increased Space Situational Awareness (SSA), and thus improved capacity for planetary defense, and thus…
- Avoidance and minimization of damage to existing and future operational space-borne systems
- The safe navigation into and through LEO, MEO, GEO and beyond; the preservation of safe space travel for future generations

It specifically offers an opportunity for space organizations share their orbital debris data in supporting and achieving these goals. Toward these goals I propose an ontological framework whose aims, in turn, are:

A. To represent the relevant knowledge and entities in the orbital debris domain
B. Foster data-sharing among orbital debris information systems (databases, space object catalogs, etc.), and perhaps form terminological and other standards in the relevant communities

The representation of orbital knowledge, data, and other entities (A) are tasks that can be accomplished within artificial intelligence, knowledge engineering, formal ontology, and *ontology development and engineering*. The idea is to explore whether ontology, as a partial approach, can contribute in solving the orbital debris problem by helping to improve SSA, in part via data-exchange and fusion. This proposal broadly calls for the creation of a data-sharing and interoperability (if not integration) system, of which ontologies can be a part, for orbital debris data, and potentially an international *joint orbital debris catalog*. Toward this, international parties are to share their orbital debris data.

**Communities and Domains of Interest**: space situational awareness, aerospace computing, astrodynamics, (astro)informatics, orbital/space debris, ontology engineering, formal ontology, applied ontology, knowledge representation and reasoning, computer science, data modeling, artificial intelligence, data science, big data, semantic web, database management.

**Minimal Project Goal:** A philosophical and conceptual analysis of orbital concepts for scientifically accurate qualitative and quantitative formal representations, including terminologies, taxonomies and ontologies. The creation of one or more ontological models (ontologies) of the orbital debris and SSA domain. In doing so, a (proposed) standardized set of category and relation terms will be presented for reuse by others in ontology, informatics, orbital debris, SSA, and astrodynamics communities.

**Project Site:** https://purl.org/space-ontology
**Orbital Debris Ontology(ODO)**: http://purl.org/space-ontology/odo

# 1. INTRODUCTION

Orbital debris is internationally recognized as a global problem. It poses a threat to persons and spacecraft in orbit, and is a potential barrier to future spaceflight. The orbital debris problem thereby presents us with an opportunity for international cooperation, and should be understood within the broader context of improving space situational awareness (SSA) for planetary defense, scientific knowledge, and preserving the future of space travel. SSA encompasses knowledge of the space environment, including that of natural and human-made objects in orbital, near-Earth and deep space regions. Knowledge of orbital debris is therefore part of SSA. Orbital debris data, then, is SSA data.

This paper focuses on one area for that cooperative potential—data sharing—and the part it plays in resolving the orbital debris problem. Originally conceived in 2011, this project concept involves researching the potential of ontology as a partial approach toward data-sharing, interoperability and knowledge discovery for the orbital debris domain. The idea is to exchange data among orbital debris and SSA communities by using (and creating) ontologies and employing formal ontological methods. Orbital debris data sharing will improve the state SSA, placing the global community in a more informed position toward orbital debris resolution. The foundational goal and motivation for this research is therefore to contribute to solving or otherwise alleviating the hazards of orbital debris, and to improve peaceful SSA. Furthermore, with greater (ideally actionable) situational knowledge of the broader space environment—near Earth to deep space regions—comes increased ability for planetary defense.

In what follows I sketch an ontological architecture for the domain of orbital debris. Specific goals of this ontological system include:

I. Orbital debris and SSA domain knowledge representation
II. Annotation of debris and SSA data
III. Orbital debris and SSA data-sharing
IV. The specification and formalization of an orbital debris terminology/vocabulary that is faithful to the respective domain knowledge and expertise.

And in so doing…

V. Contribute to: orbital debris remediation, improving SSA and space safety

Aside from the potential for computational utility vis-a-vis space informatics and space data management, a broader benefit of this project is the conceptual and philosophical analysis (and clarification) of the relevant orbital and space concepts. It is an interdisciplinary enterprise requiring the expertise and efforts from persons in astrodynamics, astroinformatics, ontology (philosophical, formal, and applied), data modeling, knowledge representation and reasoning, and computer science.

The paper is divided thusly: I summarize the problems in sections 2 and 3. Section 4 discusses ontology, and 5 outlines my orbital debris ontology. Section 6 presents the architecture with closing remarks in 7.

# 2. THE ORBITAL DEBRIS PROBLEM AND SPACE SITUATIONAL AWARENESS

**Orbital debris** is human-made, Earth-orbiting objects that are no longer usable. Examples include inoperative spacecraft, their segments; miscellaneous astronautical artifacts; as well as satellite fragments from normal operations, collisions, or explosions. The primary sources of orbital debris are collisions and explosions. The size and shape of debris vary: there are more than 21,000 objects larger than 10 cm, over 700,000 between 1 and 10 cm, and over 100 million smaller than 1 cm (NASA Orbital Debris Office). The orbital lifetime of debris varies as well: debris at 800 km altitude and debris above 1000 km remain in orbit for decades to over a century, respectively.

There are a number of hazards posed by orbital debris (Scientific American). Given their high velocities, lifetimes, and growing numbers (Figures 1, 2), there is a risk of damage to operational satellites. The probability of catastrophic events is currently low, but "[a]s the debris population grows, more collisions will occur" (European Space Agency). "[T]he continuing growth in space debris poses an increasing threat to economically and scientifically vital orbital regions" (ESA Global Experts) by increasing the spatial density of debris. This raises the likelihood of collision events and makes navigation more hazardous. Furthermore, as critical orbital regions become more congested debris pose a greater hindrance to astronomical observations. A task we are faced with is "[…] to prevent a cascade of self-sustaining collisions from setting in over the next few decades."(ESA Global Experts) The hazard of space debris also extend to Earth: debris of sufficient size pose a risk (albeit low) of harm and damage to persons and property on Earth.

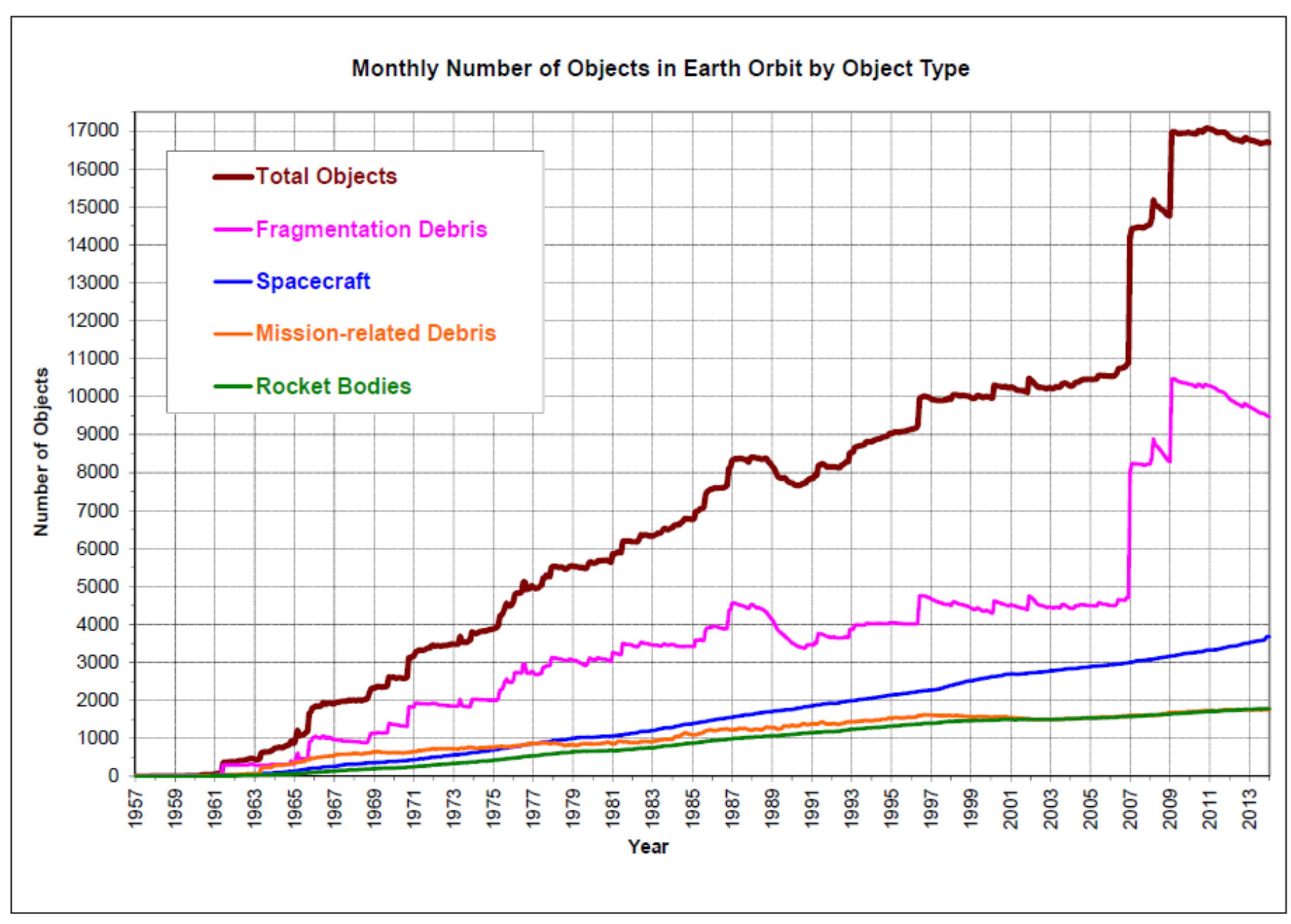

*Monthly Number of Cataloged Objects in Earth Orbit by Object Type: This chart displays a summary of all objects in Earth orbit officially cataloged by the U.S. Space Surveillance Network. "Fragmentation debris" includes satellite breakup debris and anomalous event debris, while "mission-related debris" includes all objects dispensed, separated, or released as part of the planned mission.*

**Fig.1**. Debris growth of objects ≥10cm, by type (ODQ, 2014).

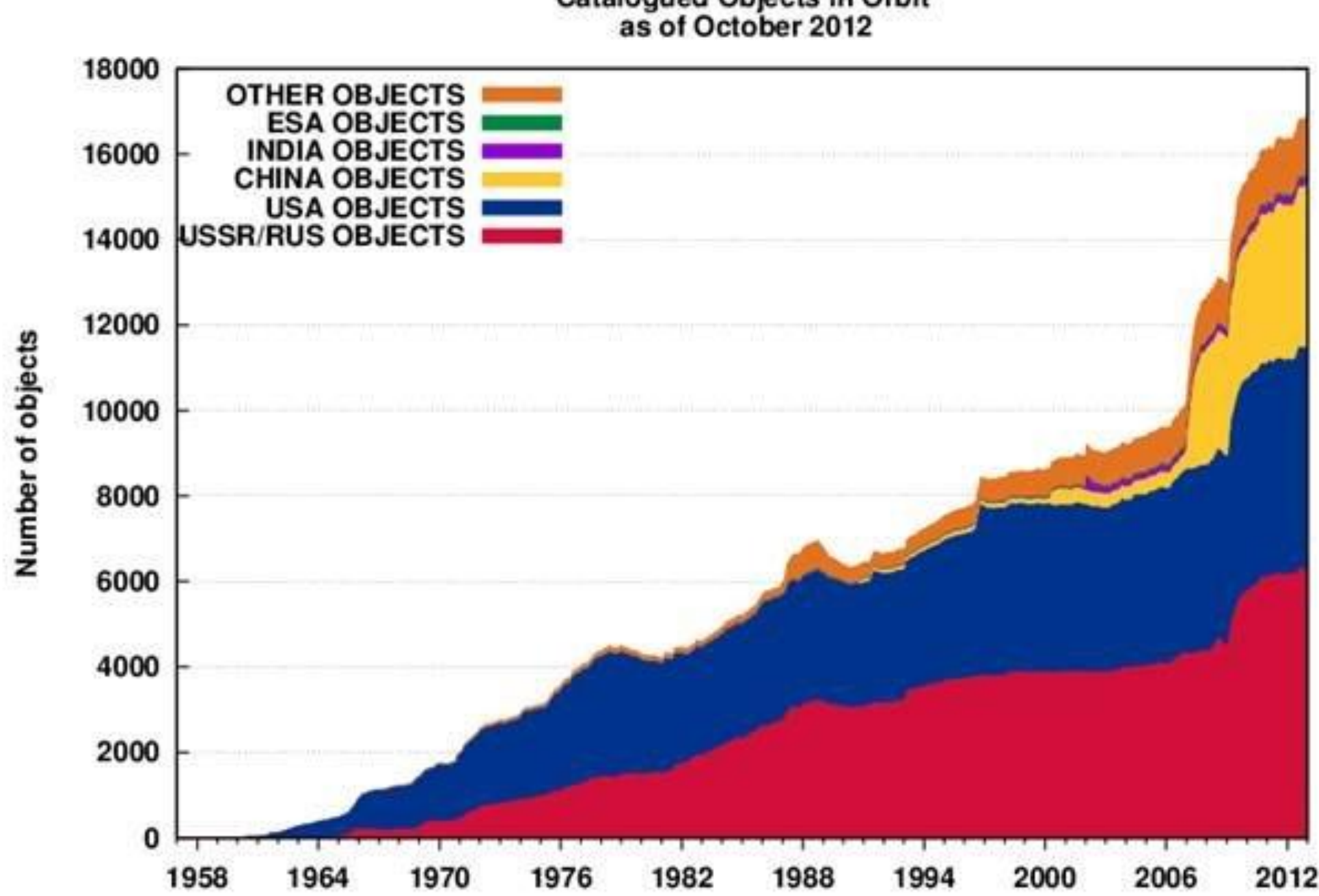

**Fig 2.** Debris growth of objects ≥10cm, by nation.[1]

Orbital debris is therefore a risk to satellites, spacecraft, human life, and has the potential to severely limit, if not prevent, future spaceflight. To the extent that debris limits access to space, and preventative actions are not taken, we are essentially trapping ourselves in. This is the **orbital debris problem**.

Resolving the orbital debris problem entails debris **prevention**, **mitigation** and **remediation** measures, the latter of which involves the development of technologies to physically remove debris. It also includes the enactment of policy that outlines requirements (Johnson and Stansbery 2010; United Nations 2010; U.S. Government), strategies, and methods to minimize the production of debris, as well as management of debris.

A necessary part of the solution to the orbital debris problem is accurate and actionable **space situational awareness**. We observe the space environment, detect, locate, identify, track, and predict debris objects and their orbital paths. Both ground-based and space-based sensors (radar and optical telescopes, etc.) are used to toward these ends. Table 1 presents a non-exhaustive list of sensors, tracking stations, sensor networks and observers that gather or process SSA data (including debris), or that maintain space object catalogs (Vallado and Griesbach 2011).

[1] European Space Agency Space Object Catalog

| Independent (public, amateur, etc.) observers[2] |
| --- |
| United States<br>• Space Surveillance Network<br>• Joint Space Operations Command<br>• Department of Defense Space Object Catalog<br>• National Aeronautics and Space Administration (NASA)<br>    o National Space Science Data Center (NSSDC)<br>    o Goldstone Deep Space Communications Complex<br>• University of Michigan Orbital Debris Survey Telescope<br>• Lincoln Space Surveillance Complex, Massachusetts Institute of Technology<br>    o Millstone Hill Radar<br>    o Haystack X-Band Radar, & Haystack Auxiliary Radar (primarily ≤10cm debris observations) (Stansbery 2009)<br>• Ground-based Electro-Optical Deep Space Surveillance<br>• Meter Class Autonomous Telescope, Ascension Island<br>• Maui Space Surveillance Complex, Air Force Maui Optical & Supercomputing Site, Maui High Performance Computing Center<br>• Space Data Center (SDC) of the Space Data Association (Private sector)<br>• Commercial Space Operations Center (CSOC) |
| United Kingdom Infrared Telescope, Hawaii |
| Russian Space Surveillance System<br>• Okno space surveillance stations |
| European Space Surveillance Sensors (ESA SST) |
| Chinese Space Surveillance System |
| International Scientific Optical Network (Geosynchronous catalog) |

**Table 1**: Sources of SSA data, and thus potential partners in data-sharing.

Any given organization tracking space debris acquires large amounts of data, yet "[n]one of the existing networks and SSA initiatives provides complete coverage or a comprehensive catalogue of all objects in orbit. **The sharing of data is also limited**." (Kretzenbacher et al 2012, emphasis added). The United States, for instance, maintains an extensive system and makes some data available, but "there are gaps in its coverage and catalog" (Weeden 2014). What more, given that it is primarily military-centric, data-sharing is not entirely at a premium (Becker and Chow 2012). The above sources of space object observational data, as well those under development and those yet to be constructed, are therefore potential partners in data-sharing initiatives toward more complete SSA. Their particular databases, space object catalogs, and information systems are candidate elements to be made interoperable via the use of ontologies or other data and knowledge management means.

The improvement of SSA for all orbital, near-earth and deep-space objects is critical for national security and **planetary defense**. In pooling data and resources toward better SSA, the individual parties expand their knowledge. Collectively they are in a better position for improving space safety and planetary defense.

> "Improving SSA for all space actors is critical to space activities and the long-term sustainability of space. It provides knowledge about what is happening in space, and in particular Earth orbit. This knowledge provides understanding of potential threats (natural or human-generated)" (Weeden 2014)

Solving the orbital debris problem and increasing SSA are more efficiently addressed via cooperative efforts at the international scale. "The removal of space debris is an environmental problem of global dimensions that must be assessed in an international context, including the UN." (ESA Global Experts) In short, "SSA is inherently international" (Weeden 2013).

[2] One or more open-access repositories for amateur observers can be formed. Such a publicly accessibly system should allow: amateur observers to enter their observations of orbital objects, and view the observations stored by others. Such a system could then share data with other parties.

# 3. THE ORBITAL DEBRIS DATA PROBLEM

Given the premise that greater data sharing is needed to improve SSA and thereby improve space safety, technical (and political) obstacles to these objectives constitute what I will call *the orbital debris* (or *SSA*) *data problem*. The supposed problem involves the state of space object catalogs, databases, or information and knowledge systems that house debris and other SSA data. In what follows, 'information system' (IS) is used as a collective term for these systems. Political obstacles aside, there are at least two specific concerns, both of which apply to other data-intensive disciplines.

The first concern is a **silo effect**: distinct, essentially isolated, information systems. Individual IS's may (i) share limited, if any, amounts of data, (ii) be inaccessible, and (iii) use different data formats, standards, terminologies, and classification schemes in spite of having a common universe of discourse. (iii) can be described as the problem of heterogeneous data and data formats, an obstacle to data-sharing and interoperability. Consider, for example, this statement expressed vis-à-vis in-orbit collision avoidance: "Unfortunately, the most accurate tracking data for active satellites is often closely held only by the satellite operators."(Kelso and Gorski 2009)

In itself a data silo is not necessarily problematic. Depending on the purpose of the IS, it may be desirable at times. For remediating the orbital debris problem, however, silos are at least potentially problematic because resolving the former entails improving SSA, and "**[f]undamentally, SSA requires data sharing and cooperation between different actors**" (Weeden 2013, emphasis added). The first concern, therefore, reflects a potential hindrance to improving the orbital debris situation. It is an indirect obstacle to ensuring safe and successful space navigation and development. At a more fundamental level, it should be easily accepted that sharing information (i.e. communication) is essential for research and scientific progress. It encourages hypothesis generation, knowledge discovery, collaboration, and division of labor.

As Kretzenbacher et al. (2012) suggest, the idea is to consolidate data for more complete coverage as well as update, if not replace, legacy systems (Weeden 2012). In sharing debris data, organizations with incomplete—and they are all incomplete—debris IS's will *increase their knowledge of existing and potential space hazards*. They will therefore have a more complete picture of the orbital environment and greater awareness toward: debris remediation, planetary defense, and assuring obstacle-free space exploration for generations to come.

We should move from situation A to either situation B or C (Figure 3), or some combination of the two. It is a move from incommunicative debris/SSA IS's to IS's sharing data with a neutral party (e.g. a consortium) formed for the creation of a joint international and dynamic master orbital debris (or SSA) catalog, and/or to interconnected and interoperable systems. In either situation, ontology may prove useful by explaining, classifying and annotating data, and serving as a partial means to interoperability.[3] The formation of a neutral party has been suggested in (Kretzenbacher et al. 2012), where we read:

> "One possible approach is the creation of a neutral international organization and network that exists solely to facilitate the collection and sharing of SSA data. The proposed network would utilize the capabilities of already existing SSA infrastructure that are not restricted by being part of a defensive network."

Weeden and Kelso (2009) also explore the idea of an international SSA system, where "the primary goal of an international civil SSA system would be to support the safe and sustainable operation of all actors through provision of necessary information" (Space Safety & Sustainability Working Group 2012, p.2). Of course, "any data sharing policy adapted by a comprehensive SSA system will need to balance privacy and security with data dissemination and openness" (Space Safety & Sustainability Working Group 2012, p.23).

[3] Interoperability, the problems thereof, and the role of ontologies, have been recently discussed in Thanos (2014).

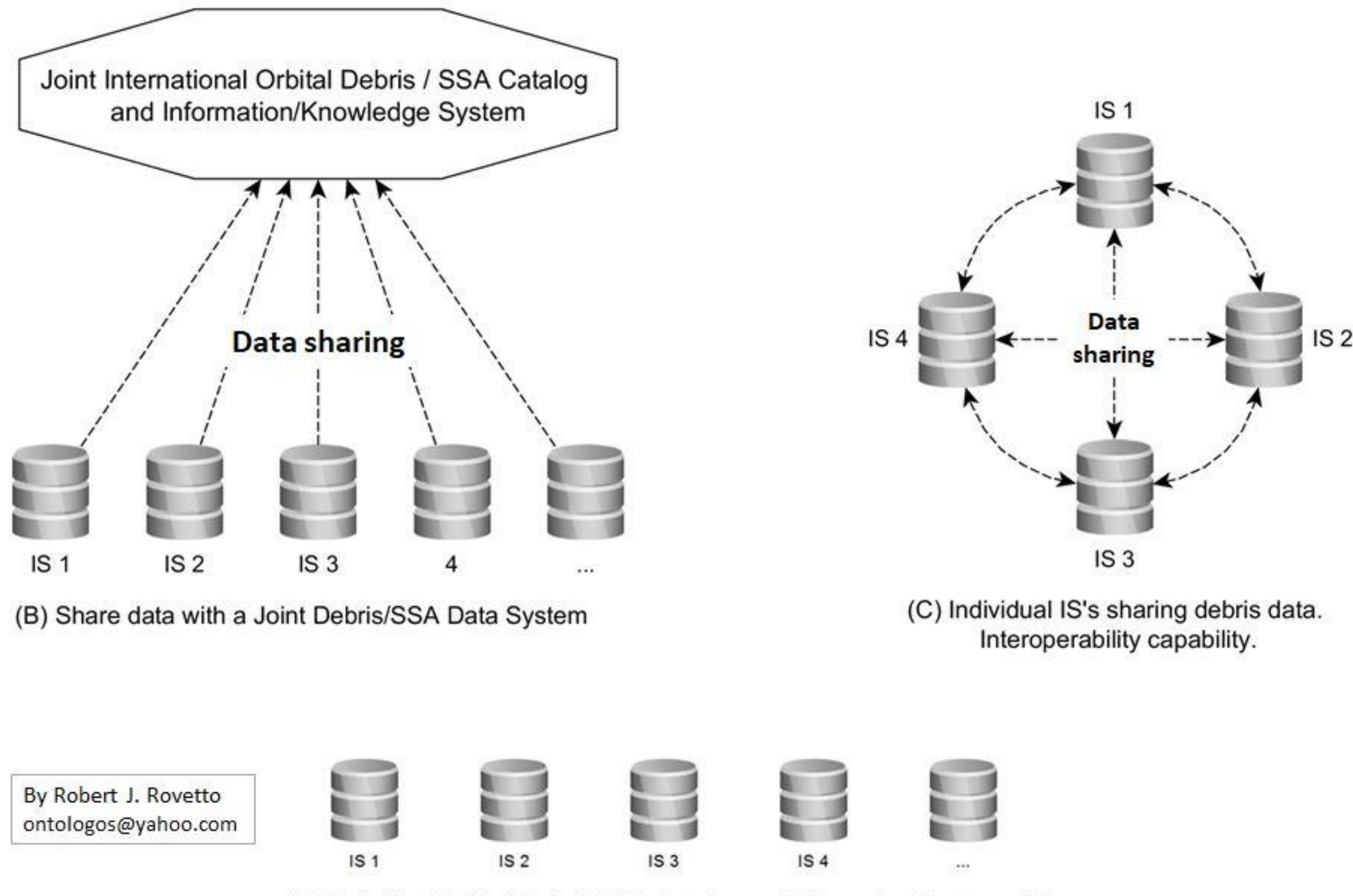

**Figure 3**: From incommunicative orbital debris catalogs to debris data-sharing. The idea is to move from (A) isolated information systems (IS), to either (B) sharing their data with a neutral joint master debris IS, or (C) share their data amongst each other in an interconnected fashion, or some combination of the two.

The second concern vis-à-vis space object catalogs and other IS's is the sheer quantity of data that is (1) acquired from sensor networks, and (2) generated from astrodynamic conjunction analysis, propagation and prediction models. To take another data-intensive field as a case in point, we read: "The progress of astronomy is about to hit a wall in terms of the processing, mining and interpretation of huge datasets." (Valentijn and Kleijn 2006, p.20) The case for advancing astroinformatics (Borne 2009) is also a case for orbital debris and SSA informatics. The orbital debris (and data) problem is thereby also an opportunity to improve the state of space informatics, and space data semantics.

Pulvermacher et al. (2000) develop an XML schema for space surveillance, applying a semantic approach to this domain along the same line of thought I propose here. For example, they use common SSA concepts, and emphasize data semantics for interoperability. In their report we also read: "maintaining a catalog of where objects are in space is a prerequisite for every other operation that uses space assets." One goal in developing one (or more) ontologies for the orbital debris domain is indeed to help maintain space object catalogs by providing a common and sharable semantically-rich terminology and classification system for those objects.

In order to improve SSA via data-sharing and interoperability, we may adopt one method used by some informaticians: ontology engineering[4]. The next section begins with a brief overview of the field of ontology, followed by a sketch of a modularized ontological architecture for representing the orbital debris and broader SSA domain.

[4] Note that I am primarily concerned with improving the orbital debris situation rather than advocating a particular method toward that goal. If ontological approaches prove to be insufficient or simply unhelpful, then there should be no quarrel with seeking alternatives.

# 4. Informatics and Ontological Engineering for Orbital Debris

Given the data-intensive nature of the orbital debris and space situational awareness domains, analyzing and communicating orbital debris datasets is necessary if we are to alleviate the orbital debris problem. The publication "Future Human Spaceflight: The Need for International Cooperation" (Pace and Reibaldi 2010) acknowledges this when it cites two data-centric areas for international cooperation:

1. "Adoption of open, interoperable communications protocols (e.g., delay-tolerant networks) defined through the Consultative Committee on Space Data Standards."

2. "Adoption of standard orbit data message formats to facilitate exchange of satellite and orbital debris location information to avoid possible collisions in space." (p.28)

"Astrodynamic standards are an important foundation for SSA" (Weeden 2012, p.23). Although the ultimate utility of this research area is still to be determined, *ontologies* provide a formalized knowledge base of a given domain, and may foster semantic interoperability via annotation and classification of data elements housed within siloed information systems.

## 4.1 Ontology for Orbital Debris

**Ontology** or **Ontological engineering** (Staab and Studer 2009) in computer science is related to knowledge representation and informatics. The ontology development process involves domain research, concept analysis, a terminology/vocabulary, and the explicit specification of at least an understanding of the given domain of interest. The terms in the taxonomy, their definitions and the interrelationships between them (or that which the terms denote), are intended to express or capture the domain. This results in a logically formalized taxonomical structure that can be computationally reasoned over.

**Ontology development** is an ongoing iterative process that requires curation, quality-control, revision and updating. This is particularly necessary for ontologies of the sciences, e.g., astronomy, where knowledge is updated and corrected, and where new discoveries are made over time. See Pinto and Martins (2004) for an account of the different stages of ontology building.

A central focus in ontology development is the *intended meaning* of the specified vocabulary. Creating **an ontology** involves the formation of the above-mentioned taxonomical structure. This hierarchical system of category and relation (predicate) terms has a formal semantics, the totality of which expresses some consensual knowledge of the domain. Some ontologies are essentially *reference ontologies*, that is, they present common and general knowledge of the given domain.

For **computer tractability**, a *logical formalization* of terms is employed using an *implementation* or *knowledge representation language*. This allows automated classification and inference, as well as querying. "Ontologies have been developed to provide a machine-processable semantics of information sources that can be communicated between different agents (software and humans)" (Fensel 2004, p.3). An ontology provides this as a classification scheme of interrelated and axiomatized categories.

In employing the commonly adopted distinction of *classes* (categories, universals, types) and their *instances* (individuals, particulars, tokens), a particular debris fragment is explicitly asserted to be an ***instance of*** the general class of **Debris**. The subclass (subtype, class subsumption) relation structures the class hierarchy. For example, the class, **Orbital Debris** is a ***subclass of*** (or simply ***Is a***) **Debris**. The former inherits the properties of the latter. Domain-neutral ontological distinctions adopted from philosophical ontology are used to help make the ontology consistent, coherent and conceptually detailed. In qualitative spatial reasoning, for instance, mereological[5] concepts are employed. Taxonomies, themselves, can be structured along partonomic lines using a ***parthood*** relation, the formal properties of which are to be asserted in the ontology.

Ontologies are intended to **minimize terminological ambiguity**. This is achieved by providing a set of general terms whose intended meanings are clearly and explicitly specified. In other words, terms are given **explicit definitions** that are comprehensible for the human user. If terms are left primitive (undefined within the ontology/system), then intuitions can provide their meaning, but some explicating description is best offered. Ontologies are minimally intended to clearly express what data is about and formally represent a shared understanding or conceptualization of a domain of interest.

Although ontologies can be reused without being composed of modules, when they are modularized, **reusability** is another benefit. "Ontologies were developed in Artificial Intelligence to facilitate knowledge sharing and reuse." (Fensel 2004, p.3) Reusing existing category and relation terms from other ontologies, for instance, is often considered desirable in ontology development.[6] Furthermore, **annotating** data elements (e.g. instance data) with ontological classes is a common task for achieving data interoperability. Axiomatized inference rules in conjunction with the generalized knowledge represented in the ontology are used to reason over that instance data.

Ontology engineering also involves **conceptual** and **philosophical analysis**, which should begin at the outset of the development process. At the most general or abstract levels, ontology development draws on ontological categories from philosophical inquiry. Examples include: Event, Object, Property, Relation, Process, State, and System. Common distinctions include: Object-Process, Object-Property, Independent-Dependent, and mereotopological concepts such as *Part, Whole*, and *Connection*. Formal ontological relations include *Participation*, *Instantiation*, *Dependence*, *Parthood*, *Constitution*, and *Causality*.

Depending on the resources at hand and the desired degree of philosophical rigor, a complete metaphysical account can be given for an ontology at all levels of generality/abstraction. For this communication I remain neutral with respect to how the above-mentioned general categories are philosophically characterized, while relying on our intuitions in the light of the scientific domain under analysis. I will, however, emphasize that the ultimate practical goals are to solve real-world problems. In this case, the overarching rationale is to contribute to spaceflight safety by improving peaceful space situational awareness via alleviating the orbital debris problem through ontology-based data sharing and integration.

# 5. Ontology Modules for Orbital Debris

This section summarizes some relevant domain ontologies for an ontology of orbital debris and SSA. Each domain ontology— ontologies of a particular subject matter or universe of discourse—should have: (i) category and relation terms, (ii) natural language definitions as well as a computable semantics, and (iii) a hierarchical structure.

## 5.1 ODO: The Orbital Debris Ontology

The **Orbital Debris Ontology**, or **ODO** for short[7], is the locus of the idea. It is an ontology of the domain or orbital debris, and as such represents the entities in, and general knowledge about, the orbital debris domain. There is one caveat. Rather than an orbital debris ontology in itself, an SSA ontology with orbital debris-relevant classes and relations may be formed instead. These classes would be portions of the SSA ontological class hierarchy. Alternatively, ODO may be part of a broader SSA ontology. In any case, desiderata for ODO, like other ontologies, include at least the following (D1-D3).

[5] The theory of parts and wholes.

[6] I would, however, argue that this depends on not only the requirements, goals of (and problems solved by) the ontological system being developed, but whether the reused (or imported) classes accurately reflect the entities in the domain in question.

[7] Pronounced: *oh-dough*

### (D1) TERMINOLOGY / VOCABULARY

Include category and relation terms specific to the domain of orbital debris. ODO will have terms for at least the following:

- Types of orbital debris
- Properties/features of debris
- Sensors observing the orbital environment (An alternative is to have a sensor ontology)
- Relations relating debris to other relevant entities
- Debris processes, including causes of debris formation
- State and Positional information[8]
  - Ephemeris data[9]
  - Orbits and orbital properties

For example, if we follow the National Research Council (1995, p.21) and the United Nations (1999), then debris types (classes) and property classes include the following (Table 2). Terms for some orbital properties include: Eccentricity, Inclination, Right Ascension of the Ascending Node, and the other Keplerian Orbital Elements/Parameters.

| Types of Debris | Properties of Debris |
|---|---|
| Non-functional Spacecraft | Mass |
| Rocket Body | Spatial Dimensions |
| Mission-related Debris | Material Composition |
| Fragmentation Debris | Radar Cross-section |
| | State Vector |
| | Albedo |
| | Ballistic Coefficient |
| | Launch Characteristics |

**Table 2**: Candidate classes for types and properties of orbital debris

Table 3 lists some category and relation terms essential to an ontology of orbital debris, and to a broader space situational awareness ontology. The domain-neutral object-process distinction reflected in the table can be formulated in various ways depending on the metaphysical accounts, but for the present communication it will be sufficient to rely on our intuitions of physical objects/things and processes/events/occurrences. Material objects, such as satellites, are formally asserted to *participate in* processes, such as orbital plane changes.

| Physical or Material Object Terms | Relation Terms | Processual Terms |
|---|---|---|
| Space Debris,<br>Orbital Debris[10] | Has_debris_source,<br>Caused_by, Has_cause | (Orbital) Collision Event,<br>Explosion Event |
| Spacecraft | Has_ephemerides | Orbital Decay Process |
| Satellite | Has_cross_section | Debris Tracking Process |
| Ground-based Sensor | Has_shape, has_diameter | Orbital Debris Mitigation Process |
| Orbit | Has_orbit, Orbits | Orbiting Process |

**Table 3**: Candidate Orbital Debris Ontology Terms.

### (D2) DEFINITIONS FOR TERMS AND A FORMAL SEMANTICS

- *Natural Language* (NL) *Definitions* that easily communicate the intended meaning to human users (comprehensibility).
- *Formal and Computable Definitions*
  First-order predicate logic, high-order or non-classical logics formalize definitions and domain knowledge. Implementation languages such as Common Logic Interchange Format (CLIF) and the Web Ontology Language (OWL) are used for computability and information exchange.

---

[8] Mentioned in task 2 from section 4.

[9] See: http://ssd.jpl.nasa.gov/?glossary&term=ephemeris

[10] Given that the class 'Space debris' can easily be interpreted as more general than 'orbital debris', I assert the latter as a type of the former. (NASA, 2008) does similarly.

- Care must be taken so as not to engage in "'over-first-orderization', i.e., trying to axiomatize too much" (Menzel 2003, p.5) and not to give inadequate axiomatizations of terms and concepts that are difficult to formalize. We must also be aware of the expressive limitations of each logic and ontology implementation language.
- Both sorts of definition should: be consistent (not self-contradictory) and convey necessary and/or sufficient conditions and *identity criteria* for each category of entity. Identity criteria include diachronic (identity across time: what makes something persist over time) and synchronic (identity at a time).

We naturally begin at the conceptual level, coming to an understanding of ideas, concepts and meanings. We then move to formulate NL definitions that express the intended meaning and conceptualizations. To the extent that we have a good understanding of the domain, our conceptualizations ideally reflect the reality of the domain. Finally, although the logical and computable definition should ideally match the NL definition as closely as possible, any disparity should be explicitly stated (and explained) in documentation. This will help avoid misinterpretations and incorrect or false inferences. Definitional revisions may take place at any point in the ontology development process.

### (D3) STRUCTURE THE TERMS, A HIERARCHICAL STRUCUTRE, A BACKBONE TAXONOMY

- Class terms should be formed into a taxonomic structure. The ***Is a*** class subsumption relation (class-subclass) is often used to structure taxonomies. This, in combination with any other relationships between classes forms a hierarchical classification scheme that affords automated classification and inheritance of properties from parent to child classes.
- Ontology editors with graphical user interfaces, such as Protégé and TopBraid Composer, make this task relatively easy to achieve. Ontology visualization tools and plugins can also improve the ontology development process, and serve as helpful pedagogical tools for users.

Achieving D1-D3 for ODO requires asking the following questions, which are part of the requisite domain research and conceptual development tasks. It is vital to consult domain literature and subject-matter/domain experts, particularly practicing domain experts, when conducting domain research and when checking the accuracy of the ontological representations, assertions, and inferences. This will help ensure high fidelity and veridicality in definitions and axiomatizations of domain knowledge. Domain experts include national space agencies such as the NASA, the European Space Agency (ESA); specific departments such as the Orbital Debris Program Office at NASA; as well as individual astrodynamicists, satellite operators, and so forth.

Q1 *How are orbital debris currently classified? How can they be classified?*
- By… size, altitude, etc.

Q2 *What types of debris are currently in orbit?*
- E.g., Spacecraft fragments, entire rocket body parts, mission-related artifacts (e.g. astronaut gloves, tools), etc.

Q3 *What are the physical, intentional and social properties of orbital debris?*
- Physical properties (relational or otherwise), include size, shape, mass, location, attitude, velocity, orbital characteristics, etc.
- Intentional and social properties include cognitive attributions, and socially constructed or legal properties and relationships, such as ownership. This also includes space policy governing the management of debris.

Q4 *What is an ontological characterization of orbital debris?*
- Answering this provides a higher-level (more general) classification and definition of 'Debris'.
  It provides one answer to the following question.

Q5 *What type of entity is orbital debris*—a Physical Object, Material Artifact, a Role, ...?'

Orbital debris may be characterized as types of **Physical Object**, i.e., spatio-temporally localized objects that have some material properties. This is intuitive because the concept (and talk) of debris brings to mind material things. However, deeper ontological analysis yields another possibility. Debris may be modeled as a **Role** we attribute to physical objects in orbit, a role that is played or borne by the object relative to social conventions, processes and systems. One might justify this alternative characterization as follows.

Common definitions describe debris as being unusable or functionless. The object in question, however, may have been functional in the past, or may have been part of a functional spacecraft. Some space objects in Earth orbit may also be reusable. As such, they cannot be described as orbital debris *at all times*. In other words, being orbital debris (or being usable/functional) is not an *essential* or *rigid property* of the object. Essential properties are those that are necessary at all times for the being the type in question. In this sense, one may wish to classify Orbital Debris as a role *played by* Physical or Material Objects in specific circumstances. Finally, and alternatively, do away with Roles and assign a socially-relative status or property to the orbiting object: functional, non-functional, usable, etc.

Many, if not most, relevant terms in this domain belong to others. 'Planet' (or 'Planetary Body'), for example, is not unique to the subject matter of orbital debris, but more accurately belongs to the discipline of astronomy. All the same a **Planet** class should be included when formalizing orbital knowledge of debris in orbit. Likewise for the **Physical Properties** of debris objects, i.e., mass, dimension, material composition, etc. Given that ODO will require terms from various disciplines it can be construed as an **application ontology** (but not necessarily so), i.e., an ontology created for a specific use or application. Application ontologies often use concepts from different domains and import terms (classes, relations) from other ontologies, including domain-specific or domain-neutral reference ontologies. The application at hand is to foster data sharing among orbital debris and space object information systems to improve the state of orbital debris SSA toward ensuring spaceflight safety.

ODO can reuse classes from ontologies of these other disciplines, if the latter are accurately formalized. Other relevant domains include: astrodynamics (orbital mechanics/dynamics) and celestial mechanics, astronomy, and satellite operations. If domain-specific ontologies for these fields do not exist, then the appropriate domain-specific classes can be asserted and defined as needed. If terms from existing ontologies are insufficient, then at least two actions can be taken. One, send a suggested curation change (with explanation) to the ontology developers in question. Two, create the desired category terms from scratch and incorporate them in ODO.

I will divide the relevant domain-specific ontologies in two groups and discuss them in the next section: Scientific Domain Ontologies and Data (or Information Object) Ontologies. Although this presupposes a distinction between information objects (e.g., data) and what the data is supposedly about, explaining the distinction is beyond the scope of this paper.

## 5.2 Scientific Domain/Reference Ontologies for ODO

These ontologies capture general scientific knowledge, including physical laws/principles of nature, of the respective discipline. Scientific domain ontologies act as reference ontologies or models expressing knowledge needed to annotate and ontologically describe instance data and ODO assertions. Given the overlap between, and the interconnectedness of, these disciplines, there will be a certain amount of prescription and arbitrariness regarding the boundaries, scope and content of each ontology. Thorough conceptual and philosophical analysis should be done for those concepts that resist placement, that are inherently vague, or that have multiple equally plausible ontological characterizations. The concept of *Orbit* may be an example.[11]

### *Astronomy Ontologies*

One or more Astronomy ontologies should contain general terms for astronomical phenomena studied by astronomers. The most conspicuous astronomical entities relevant to ODO are natural celestial bodies, i.e., planets. A broader Space Situational Awareness Ontology (SSAO) will include classes for more astronomical entities, e.g., space weather phenomena, asteroids and comets, etc.

Existing ontologies include the University of Maryland's (UMD) astronomical ontologies (Astronomy/Science Ontology), and the International Virtual Observatory Alliance (IVOA) Ontology of Astronomical Object Types. Together, this suite of ontology files includes astronomical objects, scientific units, and so forth.

---

[11] At the time of this writing I have a philosophical and formal analysis of the category of Orbit in progress.

Depending on the domain-neutral distinctions, an astronomy ontology can be divided into at least two interconnected ontologies. For example, if we assume that astronomical phenomena can be classified according to an object-process distinction, then we can form an Astronomical Object Ontology and an Astronomical Process Ontology.

The IVOA astronomical ontology includes both object and process classes in their taxonomy. The former represents what are classified as **astronomical objects**, entities such as planets, moons, stars, and comets. These entities are what are often called celestial bodies (although I use the term 'celestial' to refer to non-human made bodies in space). The latter represents planetary and stellar processes, i.e. **astronomical processes**.

Ultimately whether the object-process distinction is sufficient to capture (truthfully represent) astronomical phenomena will depend on how it is specified, i.e. on the definitions of Object and Process classes. In general, it is not clear that all astronomical phenomena can, in fact, be faithfully categorized as either object or process in the traditional philosophical sense. A hybrid of the two classes, such that objects have processual properties and processes have object-like properties, might be in order for instance. From a philosophical perspective, cases can be made for various metaphysical characterizations, but this is an area for further discussion elsewhere. I would suggest that employing a richer ontological distinction, such as **object-process-system** may be helpful.

## *CELESTIAL MECHANICS & ASTRODYNAMICS ONTOLOGIES*

An Astrodynamics Ontology represents knowledge of astrodynamics, i.e., the motion of *human-made* objects in space. This is a primary focus for ODO because we are interested in the current and future orbital states (position, velocity, etc.) of debris. Candidate category terms include those for the **Keplerian Orbital Elements/Parameters**, **spacecraft maneuvers**, and the **types of spacecraft orbits**. Instance data includes data about particular maneuvers, and the particular orbital elements defining the orbits of individual spacecraft, debris objects and other space objects at one or more times.

Given the shared scientific foundations, astrodynamics is closely related to celestial mechanics. A Celestial Mechanics Ontology—to the extent it is distinct from an Astrodynamics Ontology—is of the orbital motion of *natural* bodies in space. Due to the shared fundamentals, these two fields may be expressed in a single ontology.

Although both classical and modern relativistic physics should be taken into account, whether both are in fact expressed in the ontologies may ultimately depend on the purpose of the system and the resources at hand. Classical physics is retained for its wide applicability, but the Relativistic physics must be included not only because it is more precise, but because successful satellite operations require it.

Candidate classes for these ontologies include: **Orbital Velocity**, **Orbital Pertubation**, **Astronomical Coordinate System**, and **Keplerian Orbital Element(s)/Parameter(s)**, which includes **Inclination**, **Eccentricity**, **Right Ascension of the Ascending Node**, and the other elements. Instances of these categories include specific velocities, and particular perturbing forces or processes affecting the motion and orbit of a spacecraft.

Toward a taxonomy of these disciplines, the category **Space Dynamics**, may subsume that of **Astrodynamics** and **Celestial Dynamics**. The latter two preserve the artificial-natural connotations while the former is added to be more general than either of the two.

The development of these scientific domain ontologies involves analyzing and ontologically representing the data formats for encoding the orbital elements, formats such as the Two-line Element Set (TLE). We must determine whether classes for each orbital element are best placed within one of these ontologies, or in an information/data/measurement or geometric object ontology. Given the extensive use of physical-geometric concepts, a combination of terms representing physical phenomena, measurements and geometric constructs will be involved. This research topic provides interesting investigations into: the ontology of physical geometry; philosophy of space and time; the relationship between geometric/mathematical entities and physical phenomena; and the philosophy of science.

## 5.3 Data Type, Information & Geometric Entity Ontologies

Data/Information Ontologies and Geometric Ontologies represent the data types, standard units, data formats, information objects (or entity) and geometric objects. Coordinate systems, conceived as a geometric construct, are to be represented in these ontologies. Note that the concept of a **data** or **information entity** is not entirely clear and free of problems, but it will suffice to make the point. It is also unclear whether geometric entities are correctly classified as a type of these entities, types of abstracta, cognitive artifact, or otherwise.

Agencies such as NASA and National Institute of Standards and Technology (NIST) are notable sources for this knowledge. Formal representations of these sorts of entity exist as well. The IVOA and UMD ontologies, for example, have measurement and unit portions, and the NASA Semantic Web for Earth Environmental Terminology (SWEET) ontologies include a number of terms for units. Finally, the Quantities, Units, Dimensions and Data Types Ontologies (QUDT) also offer data type ontologies.

### *Astrodynamics Standards Ontology*

An astrodynamics standards ontology should include standard formats for representing coordinate systems, and ephemeris data among other entities. The NORAD Two-Line Element Set (NASA TLE) is the most prominent standard format for defining an orbit. Orbital State Vectors, by contrast, is another format used to express the position of an object in orbit. The referents (if there are any) of each data element—presumably the Keplerian Orbital Elements—in the TLE should be determined. For example, Orbital Inclination as an angle between a reference plane (a geometric construction) and the orbital plane, is apparently an Angle (an information, mathematical or geometric entity). More generally, it may be characterized as a Relational Property. For SSA organizations with distinct formats for the same orbital elements, an ontology serves to provide common terminology annotating and classifying each common orbital parameter.

Table 4 provides an example domain-specific category terms for the ontologies just discussed. For information on astrodynamic standards see Vallado (2001).

| **Domain Ontology** | Astronomy Ontologies | Astrodynamics Ontology | Astrodynamics Standards/Data Format Ontology |
|---|---|---|---|
| **Category Term** | Celestial Body, Planetary Body, Planet, Planetary System, Precession, Planetary Orbit, Planetary Orbiting Process, … | Ephemeris | Two-line Element Set |

**Table 4**: Example general terms (classes) for relevant scientific domain ontologies

## 5.4 An Example Debris Scenario

When identifying and tracking an object in space, data is gathered and extrapolated on a variety of features of the object and its motion, such as size, shape, attitude, velocity, direction, position. This applies to any space object in orbit: from operational satellites to orbital debris. The classical orbital parameters may be treated as properties of object's motion, its orbit, or the object itself, depending on further analysis. These Keplerian parameters describe the object's orbit, are determined at a time (an epoch), and propagated forward in time to yield a predicted orbital path. As such, this research presents a potential case study for investigating the relationship of predictive models/modeling to scientific inquiry and ontology.

Toward ontologically representing orbital scenarios, I will use the February 10, 2009 collision of the Iridium 33 and Cosmos 2251 satellites (Orbital Debris Quarterly 2009)[12] as an example. Consider this arbitrary natural language expression describing the scenario. Far more is to be said for any orbital situation, but this will suffice to make the point.

[12] See also: http://swfound.org/media/6575/swf_iridium_cosmos_collision_fact_sheet_updated_2012.pdf, and http://celestrak.com/events/collision/

> Debris fragment 1993-036BLP was formed by the collision of Irridium 33 and Cosmos 2251 satellites at time t. 1993-036BLP has a near circular orbit at an average altitude of 934 km, with Two-line Element set [E1,…,E6][13] for time t (epoch).

The expression refers to a particular Orbital Debris Object (1993-036BLP), a past Event (the collision), Properties of the debris orbit (and/or the debris itself), and Measurements. '1993-036BLP' is the international designator (name, label, etc.) denoting an individual debris fragment formed from the collision, but I do not include particular orbital parameters. Rather, I use '[E1…E6]' as a general placeholder to denote the specific TLE set expressing those six temporally-indexed parameters. The collision of the two satellites is represented by 'Coll-IC'.

Table 5 lists terms for individuals and undefined classes and relations for axiomatizing the scenario. More than one possible predicate are included to express relations between debris and their genesis.

| **Individuals / Instances** | **General Classes** | **Relations** |
|---|---|---|
| Earth | Planet | Instantiation (*instance_of*) |
| Irridium 33 | Satellite | Subsumption (*Is a*) |
| Cosmos 2251 | Communications Satellite | Causation (*has_cause*)<br>has_formation_event |
| Coll-IC | Satellite Collision Event | has_orbit |
| 1993-036BLP | Orbital Debris<br>Orbital Debris Fragment | has_altitude |
| | Orbit, Orbiting Process | orbits |

**Table 5**: Individuals, classes and relations for ontologically characterizing the satellite collision scenario

Below is a set of formal assertions using a common prefix notation to express the orbital scenario in a computable manner. The assertion '*is_an_instance_of*(Earth, **Planet**)', for example, is read as 'Earth is an instance of **Planet**'. Class terms are in bold, relation terms (predicates) are italicized, and instances are in plain text. Compare these with table 6, which presents the assertions in a subject-predicate-object form similar to the Resource Description Framework (RDF). Each row is read from left to right and expresses some specific fact of the orbital scenario. Although the formal assertions have only two arguments, n-ary predicates will be required for increased expressivity, i.e., to better capture the real-world scenarios and general knowledge of the orbital domain. For this reason, more expressive knowledge representation languages, such as CLIF, are preferable.

| **Subject** | **Predicate** | **Object** |
|---|---|---|
| Earth | *is_an_instance_of* | Planet |
| 1993-036BLP | *is_an_instance_of* | Orbital Debris (Fragment) |
| 1993-036BLP | *has_label, has_name,*<br>*has_international_designator* | "1993-036BLP" |
| 1993-036BLP | *Orbits* | Earth |
| 1993-036BLP | *has_formation_event*,<br>*was_formed_by, has_cause* | coll-IC |
| coll-IC | *is_an_instance_of* | Satellite Collision Event |

**Table 6**: Subject-Predicate-Object form of formal assertions (read left to right for each row)

*is_an_instance_of* (1993-036BLP, **Orbital Debris Fragment**)
*is_an_instance_of* (coll-IC, **Satellite Collision Event**)

*has_cause* (1993-036BLP, coll-IC)
*has_formation_event* (1993-036BLP, coll-IC)

*orbits* (1993-036BLP, Earth)

*has_altitude* (1993-036BLP, 934 km)

[13] Each data element in the TLE can be decomposed into sub-parts as needed.

*has_label* (1993-036BLP, "1993-036BLP")
*has_international_designator* (1993-036BLP, "1993-036BLP")

The term for altitude, like other quantities, can be decomposed into a numerical value and a distance unit. Furthermore, if we wish to explicitly mention the orbit as a particular entity unto itself (albeit a relational one) with orbital properties, we may assert: *has_orbit*(1993-036BLP, O1). We can then describe O1 (the orbit of the debris) and its properties (the Classical Orbital Parameters) by relating it to some TLE set or Orbital State Vector. Each Orbital Parameter—Inclination, Right Ascension of the Ascending Node, Eccentricity, etc.—should have its own corresponding class term, instances of which reflect the orbital properties of some space object in orbit. TLE sets (or otherwise) include terms denoting the measured values of these parameters. If 'TLE-1' is asserted to refer to some specific TLE set of 1993-036BLP, we may then assert the following expressions:

*is_an_instance_of* (O1, **Orbit**)
*is_described_by* (O1, TLE-1)

Another possibility to model orbital characteristics is to describe the debris object as participating in an orbiting *process* (or orbital/orbit process) that is described by a TLE set at a time, or that has specific orbital parameters as properties (at a time). However we represent these entities, ODO, or any space object or space situational awareness ontology, will require capturing general rules and knowledge about orbits. For any orbital object (debris or otherwise), for example, the formal statement '*has_orbit* (**Orbital Debris Object**, **Orbit**)', which expresses the idea that all debris objects have or are in some orbit, appears to hold true. In other words, if an object is an orbital debris object, then it is in orbit. Furthermore, all orbits have (or are described by) some orbital parameters (inclination, eccentricity, etc.). More specifically, if x is an instance of **Orbit**, then x *has orbital parameter* **Inclination** (and so on).

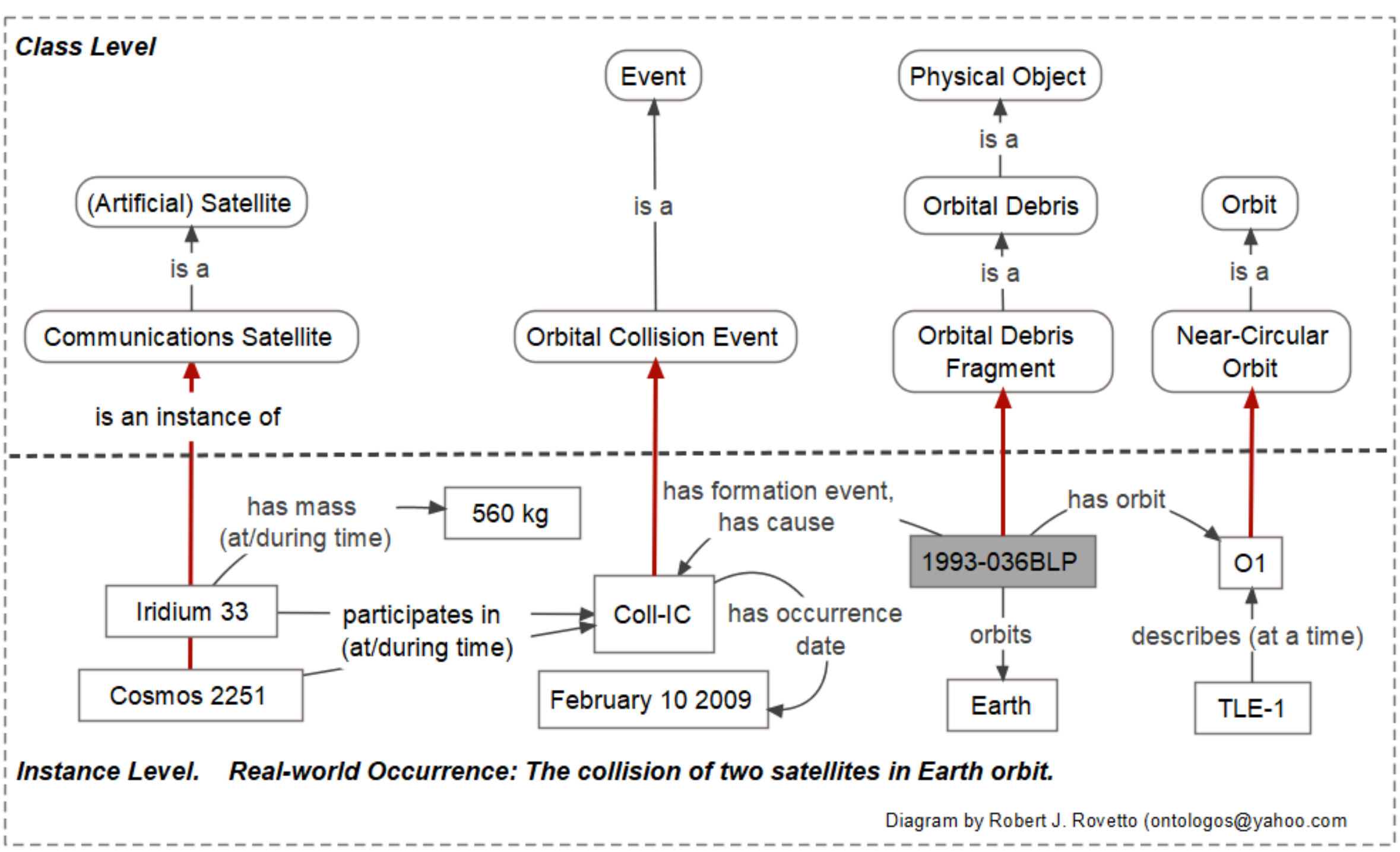

**Figure 4.** A diagram of the ontological characterization of an orbital event in which two communication satellites—Iridium 33 and Cosmos 2251—collided in Earth orbit, forming various debris fragments. The upper portion represents the class- or type-level, with the lower portion representing the instance or individual level (the particular event, and physical objects in question). Red arrows signify the instantiation relation. 'Is a' signifies the class-subclass (subsumption) relation. The grey box represents an individual orbital debris fragment formed by the collision (Coll-IC).

The instance data—the particular values—of the TLE for this debris fragment are to be annotated with their corresponding class terms, e.g. **Inclination**, **Eccentricity**, **Right Ascension of the Ascending Node**, etc. Finally, some of the above relations may be ternary, rather than binary, where the third argument is a time instant or interval.

# 6. AN ARCHITECTURE FOR AN ONTOLOGICAL FRAMEWORK FOR ORBITAL DEBRIS

Figure 4 presents a unified ontological framework for the domain of orbital debris (or SSA more broadly). It is also interpretable as portraying the scenarios from Figure 3. The green rectangular figure represents individual space object catalogs (orbital debris information systems) housing instance data about particular defunct satellites, spacecraft and other space objects considered as orbital debris. Instance data, which includes sensor data of actual space objects, should be annotated with ODO general categories to provide meaning and a formal semantics to the user and computational system, respectively. ODO (re)uses terms from scientific, engineering and data type (or information entity) ontologies (if they exist). Domain-specific ontologies may be subsumed by more general ontologies and a top-level ontology, the latter of which consists of domain-neutral categories and relations. This framework is offered as one possibility, subject to change, and makes no claims to completeness.

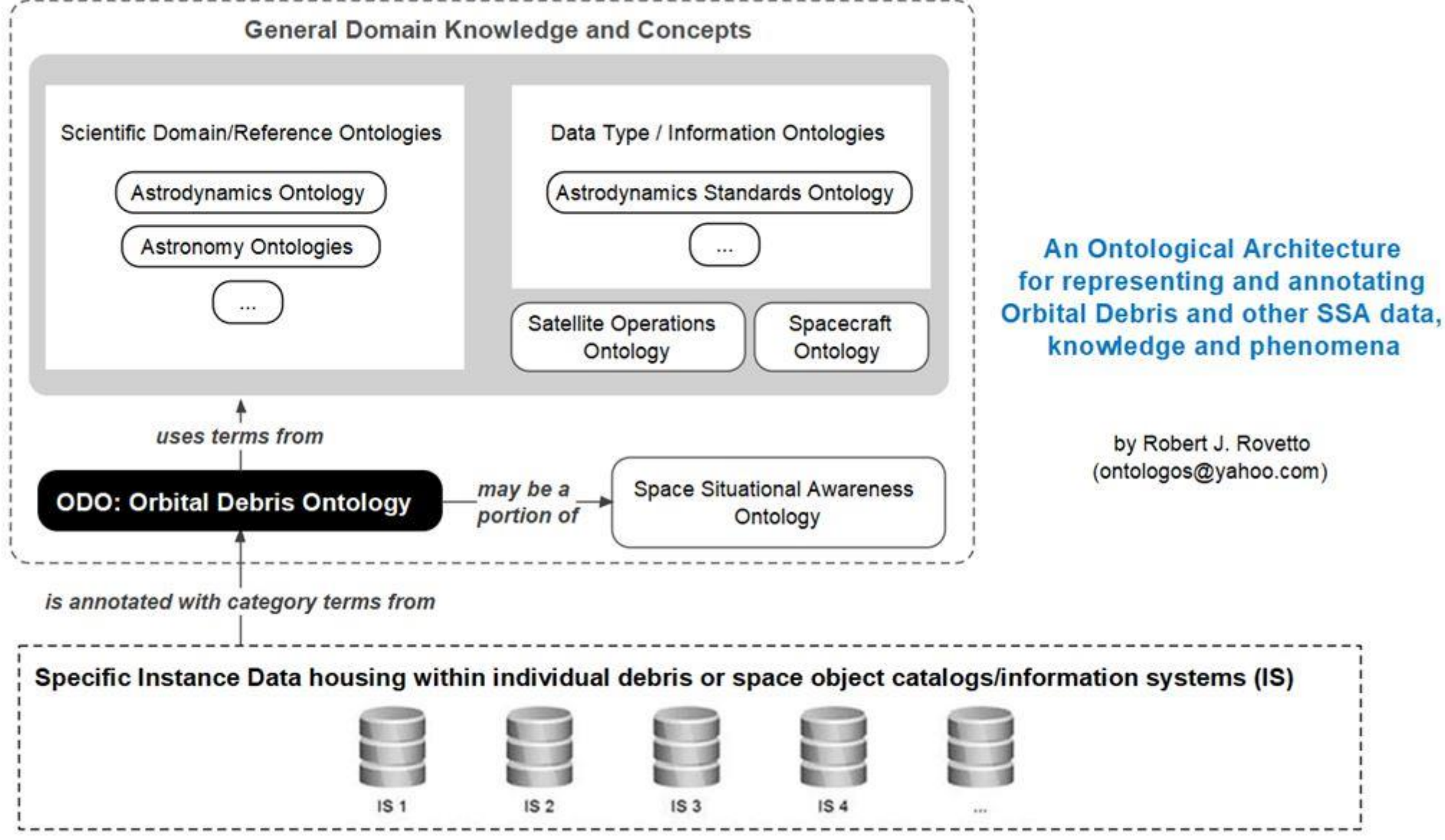

**Figure 5.** An ontological architecture for Orbital Debris and other SSA data. Lower figure represents disparate SSA or debris information systems. Examples may include those from the organizations from table 1. The relevant data and knowledge is classified, characterized by, and annotated with, terms from ODO (or a broader SSA ontology), which uses terms from the relevant interconnected domain ontologies.

In summation, individual orbital debris (or SSA) object information and systems are to be interconnected or otherwise share debris data toward a more exhaustive coverage of debris SSA. The application of ontologies may foster this by providing a common and formally specified orbital debris and space situational awareness terminology and knowledge base. Category terms from ODO may be a portion or subset of a broader SSA ontology. Terms from more general, yet domain-specific scientific and engineering disciplines, will be incorporated. In creating precise definitions of domain-specific terms and concepts, this research may contribute to the development of improved astrodynamic and SSA standards toward advancing the state of space development.

Toward implementation of this ontological architecture, the following suggestions are made. First, ontology development tools such as Protégé, or TopBraid Composer, will helpful in the process. I have, for instance, created an ODO/SSA ontology file using Protégé, adding essential orbital classes. Second, I suggest the implementation language be as expressible as possible. For this reason, at least full first-order knowledge representation languages are recommended. CLIF, for instance, is preferable to OWL despite the latter being more widespread. Finally, given the global nature of the real-world problem, this pursuit should have national space agencies as partners and participants, agencies such as the European Space Agency (ESA), NASA, and the Russian Federal Space Agency, among other burgeoning space agencies from around the world.

# 7. Conclusion

By improving our situational awareness of the orbital environment, we are in a better position to resolve the threat of space debris, and improve spaceflight safety. Given (i) the existence of disparate space situational awareness information systems from various space agencies, (ii) heterogeneous data and data formats, (iii) the gaps in orbital coverage, and therefore (iv) our limited knowledge of the orbital debris and broader space object population, orbital debris data sharing is intended to increase peaceful global SSA.

Toward these goals, I have presented an ontological architecture for the orbital debris domain and outlined some requirements for an orbital debris ontology (ODO). The purpose of ODO (or a broader SSA ontology) is to represent general knowledge of, and the entities within, the orbital debris and SSA domain; offer a common orbital debris lexicon; and foster data-sharing among space debris/object catalogs and SSA information systems. Further work includes: additional domain research; conceptual and philosophical analysis of orbital and astrodynamic concepts; an ontological description of the Two-line Element set; a first- or higher-order formalization of the relevant classes, and a computable specification.

More generally, I hope this paper has presented ideas for advancing and improving space situational awareness via space data exchange, integration, and the ontology engineering for astronautics.